%% file: acl2021.tex
\documentclass[11pt,a4paper]{article}
\usepackage[hyperref]{acl2021}
\usepackage{times}
\usepackage{latexsym}

\usepackage{soul}
\usepackage{url}
\usepackage{float}
\usepackage[utf8]{inputenc}
\usepackage{graphicx}
\usepackage{amsmath}
\usepackage{amssymb}
\usepackage{booktabs}
\usepackage{multirow}
\usepackage{array}
\newcolumntype{P}[1]{>{\centering\arraybackslash}p{#1}}

\usepackage[hang]{footmisc}
\usepackage{microtype}

\aclfinalcopy 
\def\aclpaperid{1804} 

\title{VisualSparta: An Embarrassingly Simple Approach to Large-scale Text-to-Image Search with Weighted Bag-of-words}

\author{Xiaopeng Lu\thanks{This work was partially done during an internship at SOCO} \\
  Language Technologies Institute\\
  Carnegie Mellon University \\
  \texttt{xiaopen2@andrew.cmu.edu} \\\And
  Tiancheng Zhao, Kyusong Lee\\
  SOCO Inc \\ 
  Pittsburgh, USA \\
  \texttt{\{tianchez,kyusongl\}@soco.ai} \\}

\date{}

\begin{document}

\maketitle

\begin{abstract}
Text-to-image retrieval is an essential task in cross-modal information retrieval, i.e., retrieving relevant images from a large and unlabelled dataset given textual queries. In this paper, we propose VisualSparta, a novel (\textbf{Visual}-text \textbf{Spar}se \textbf{T}ransformer M\textbf{a}tching) model that shows significant improvement in terms of both accuracy and efficiency. VisualSparta is capable of outperforming previous state-of-the-art scalable methods in MSCOCO and Flickr30K. We also show that it achieves substantial retrieving speed advantages, i.e., for a 1 million image index, VisualSparta using CPU gets $\sim$391X speedup compared to CPU vector search and $\sim$5.4X speedup compared to vector search with GPU acceleration. Experiments show that this speed advantage even gets bigger for larger datasets because VisualSparta can be efficiently implemented as an inverted index. To the best of our knowledge, VisualSparta is the first transformer-based text-to-image retrieval model that can achieve real-time searching for large-scale datasets, with significant accuracy improvement compared to previous state-of-the-art methods.

\end{abstract}

\section{Introduction}
\input{intro}

\section{Related Work}
\input{related_work}

\section{VisualSparta Retriever}
\input{method}

\section{Experiments}
\input{experiments}

\section{Model Analysis}
\input{analysis}

\section{Conclusion}
\input{conclusion}

\bibliographystyle{acl_natbib}
\bibliography{acl2021}
\end{document}

%% file: intro.tex
Text-to-image retrieval is the task of retrieving a list of relevant images from a corpus given text queries. This task is challenging because in order to find the most relevant images given text query, the model needs to not only have good representations for both textual and visual modalities, but also capture the fine-grained interaction between them. 

\begin{figure}[ht]
\centering
\includegraphics[width=0.48\textwidth]{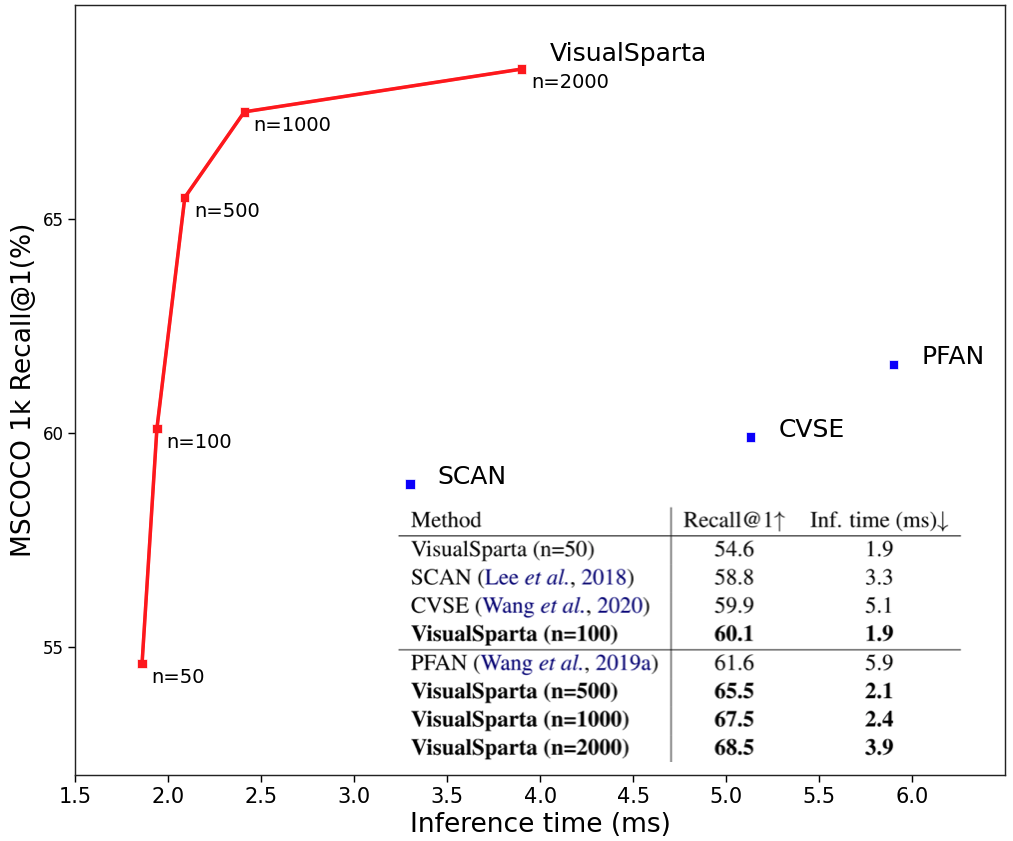}
\caption{\textbf{Inference Time vs. Model Accuracy.} Each dot represents Recall@1 for different models on MSCOCO 1K split. By setting top \textit{n}-terms to 500, our model significantly outperforms the previous best query-agnostic retrieval models, with $\sim$2.8X speedup. See section \ref{sec:speed-accu-flex} for details.} 
\label{fig:visualsparta-r1}
\end{figure}

Existing text-to-image retrieval models can be broadly divided into two categories: query-agnostic and query-dependent models. The dual-encoder architecture is a common query-agnostic model, which uses two encoders to encode the query and images separately and then compute the relevancy via inner product~\cite{faghri2017vse++,lee2018stacked,wang2019position}. The transformer architecture is a well-known query-dependent model~\cite{devlin2018bert,yang2019xlnet}. In this case, each pair of text and image is encoded by concatenating and passing into one single network, instead of being encoded by two separate encoders~\cite{lu202012,li2020oscar}. This method borrows the knowledge from large pre-trained transformer models and shows much better accuracy compared to dual-encoder methods~\cite{li2020oscar}.


Besides improving the accuracy, retrieval speed has also been a long-existing subject of study in the information retrieval (IR) community~\cite{manning2008introduction}. Query-dependent models are prohibitively slow to apply to the entire image corpus because it needs to recompute for every different query. On the other hand, query-agnostic model is able to scale by pre-computing an image data index. For dual-encoder systems, further speed improvement can be obtained via Approximate Nearest Neighbors (ANN) Search and GPU acceleration~\cite{johnson2019billion}.

In this work, we propose VisualSparta, a simple yet effective text-to-image retrieval model that outperforms all existing query-agnostic retrieval models in both accuracy and speed. By modeling fine-grained interaction between visual regions with query text tokens, our model is able to harness the power of large pre-trained visual-text models and scale to very large datasets with real-time response. To our best knowledge, this is the first model that integrates the power of transformer models with real-time searching, showing that large pre-trained models can be used in a way with significantly less amount of memory and computing time. Lastly, our method is embarrassingly simple because its image representation is essentially a weighted bag-of-words, and can be indexed in a standard Inverted Index for fast retrieval. Comparing to other sophisticated models with distributed vector representations, our method does not depend on ANN or GPU acceleration to scale up to very large datasets.

Contributions of this paper can be concluded as the following: 
\textbf{(1)} A novel retrieval model that achieves new state-of-the-art results on two benchmark datasets, i.e., MSCOCO and Flickr 30K. 
\textbf{(2)} Weighted bag-of-words is shown to be an effective representation for cross-modal retrieval that can be efficiently indexed in an Inverted Index for fast retrieval. 
\textbf{(3)} Detailed analysis and ablation study that show advantages of the proposed method and interesting properties that shine light for future research directions.

%% file: related_work.tex
Large amounts of work have been done on learning a joint representation between texts and images~\cite{karpathy2015deep,huang2018learning,lee2018stacked,wehrmann2019language,li2020oscar,lu202012}. In this section, we revisit dual-encoder based retrieval model and transformer-based retrieval model.

\subsection{Dual-encoder Matching Network}
Most of the work in text-to-image retrieval task choose to use the dual-encoder network to encode information from text and image modalities. In~\citet{karpathy2015deep}, the author used a Bi-directional Recurrent Neural Network (BRNN) to encode the textual information and used a Region Convolutional Neural Network (RCNN) to encode the image information, and the final similarity score is computed via the interaction of features from two encoders. \citet{lee2018stacked} proposed stacked cross-attention network, where the text features are passed through two attention layers to learn interactions with the image region. \citet{wang2019position} encoded the location information as yet another feature and used both deep RCNN features~\cite{ren2016faster} and the fine-grained location features for the Region of Interest (ROI) as image representation. In \citet{wang2020consensus}, the author utilized the information from Wikipedia as an external corpus to construct a Graph Neural Network (GNN) to help model the relationships across objects. 

\subsection{Pre-trained Language Models (PLM)}
Large pre-trained language models (PLM) show great success over multiple tasks in NLP areas in recent years~\cite{devlin2018bert,yang2019xlnet,dai2019transformer}.  After that, research has also been done on cross-modal transformer-based models and proves that the self-attention mechanism also helps jointly capture visual-text relationships~\cite{li2019visualbert,lu202012,qi2020imagebert,li2020oscar}. By first pretraining model under large-scale visual-text dataset, these transformer-based models capture rich semantic information from both texts and images. Models are then fine-tuned for the text-to-image retrieval task and show improvements by a large margin. However, the problem of using transformer-based models is that it is prohibitively slow in the retrieval context: the model needs to compute pair-wise similarity scores between all queries and answers, making it almost impossible to use the model in any real-world scenarios. Our proposed method borrows the power of large pre-trained models while reducing the inference time by orders of magnitude.

PLM has shown promising results in Information Retrieval (IR), despite its slow speed due to the complex model structure. The IR community recently started working on empowering the classical full-text retrieval methods with contextualized information from PLMs~\cite{dai2019context,macavaney2020expansion,zhao2020sparta}. ~\citet{dai2019context} proposed DeepCT, a model that learns to generate the query importance score from the contextualized representation of large transformer-based models. ~\citet{zhao2020sparta} proposed sparse transformer matching model (SPARTA), where the model learns term-level interaction between query and text answers and generates weighted term representations for answers during index time. Our work is motivated by works in this direction and extends the scope to the cross-modal understanding and retrieval.


%% file: method.tex
In this section, we present VisualSparta retriever, a fragment-level transformer-based model for efficient text-image matching. The focus of our proposed model is two-fold: 
\begin{itemize}
    \item Recall performance: fine-grained relationship between queries and image regions are learned to enrich the cross-modal understanding.
    \item Speed performance: query embeddings are non-contextualized, which allows the model to put most of the computation offline. 
\end{itemize}

\begin{figure*}[ht]
\centering
\includegraphics[width=14cm]{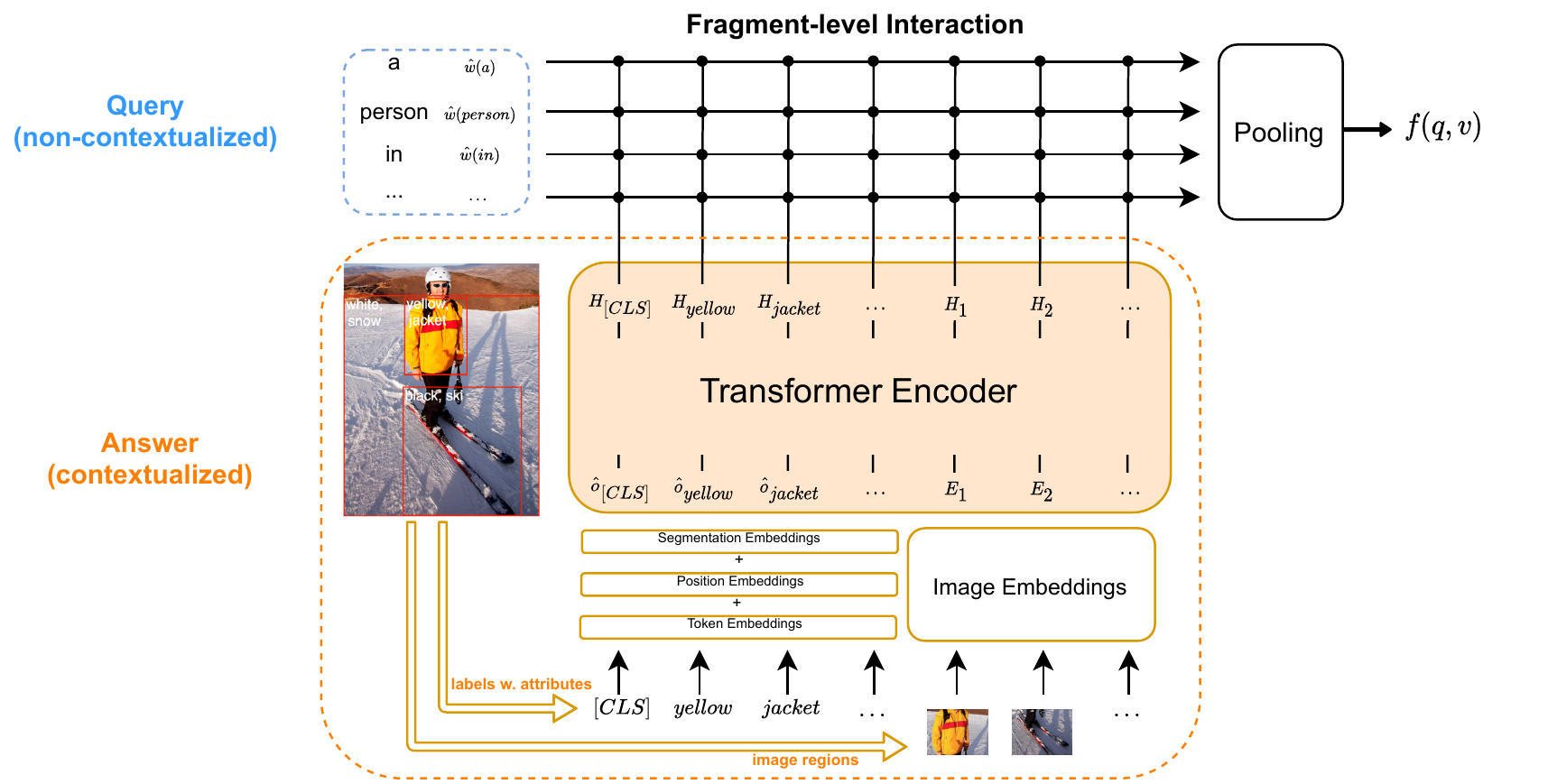}
\caption{VisualSparta Model. It first computes contextualized image region representation and non-contextualized query token representation. Then it computes a matching score between every query token and image region that can be stored in an inverted index for efficient searching.}
\label{fig:visualsparta-model}
\end{figure*}

\subsection{Model Architecture}
\subsubsection{Query representation}
\label{sec:query-rep}
As query processing is an online operation during retrieval, the efficiency of encoding query needs to be well considered. Previous methods pass the query sentence into a bi-RNN to give token representation provided surrounding tokens~\cite{lee2018stacked,wang2019position,wang2020consensus}. 

Instead of encoding the query in a sequential manner, we drop the order information of the query and only use the pretrained token embeddings to represent each token. In other words, we do not encode the local contextual information for the query and purely rely on independent word embedding $E_{tok}$ of each token. Let a query be $q = [w_1, ..., w_m]$ after tokenization, we have:
\begin{align}
    \hat{w}_{i}=E_{tok}\left(w_{i}\right)&
\end{align}
where $w_{i}$ is the $i$-th token of the query. Therefore, a query is represented as 
$\hat{w} = \{\hat{w}_{1}, ... , \hat{w}_{m}\}, \hat{w}_{i}\in \mathbb{R}^{d_{H}}$. In this way, each token is represented independently and agnostic to its local context. This is essential for the efficient indexing and inference, as described next in section \ref{method:eff-index}.

\subsubsection{Visual Representation}
Compared with query information which needs to be processed in real-time, answer processing can be rich and complex, as answer corpus can be indexed offline before the query comes. Therefore, we follow the recent works in Vision-Language Transformers~\cite{li2019visualbert,li2020oscar} and use the contextualized representation for the answer corpus.

Specifically, for an image, we represent it using information from three sources: regional visual features, regional location features, and label features with attributes, as shown in Figure~\ref{fig:visualsparta-model}. 

\paragraph{Regional visual features and location features} 
Given an image $v$, we pass it through Faster-RCNN~\cite{ren2016faster} to get $n$ regional visual features $v_i$ and their corresponding location features $l_i$:
\begin{equation}
v_1, ..., v_n=\operatorname{RCNN}(v), v_i \in \mathbb{R}^{d_{rcnn}}
\end{equation}
and the location features are the normalized top left and bottom right positions of the region proposed from Faster-RCNN, together with the region width and height:
\begin{equation}
l_i=[l_{xmin}, l_{xmax}, l_{ymin}, l_{ymax}, l_{width}, l_{height}]\\    
\end{equation}
Therefore, we represent one region by the concatenation of two features:
\begin{gather}
E_i = [v_i; l_i]\\
E_{image}=[E_1,...,E_n], E_i\in \mathbb{R}^{d_{rcnn}+d_{loc}}
\end{gather}
where $E_{image}$ is the representation for a single image. 

\paragraph{Label features with attributes}
Additional to the deep representations from the proposed image region, previous work by~\citet{li2020oscar} shows that the object label information is also useful as an additional representation for the image. We also encode the predicted objects and corresponding attributes obtained from Faster-RCNN model with pretrained word embeddings:
\begin{gather}
    \hat{o_i} = E_{tok}(o_i)+E_{pos}(o_i)+E_{seg}(o_i)\\
    E_{label} = [\hat{o_1},...,\hat{o_k}], \hat{o_i} \in \mathbb{R}^{d_{H}}
\end{gather}
where k represents the number of tokens after the tokenization of attributes and object labels for $n$ image regions. $E_{tok}$, $E_{pos}$, and $E_{seg}$ represent token embeddings, position embeddings, and segmentation embeddings respectively, similar to the embedding structure in ~\citet{devlin2018bert}. 


Therefore, one image can be represented by the linear transformed image features concatenated with label features:
\begin{equation}
    a = [(E_{image}W+b); E_{label}]
\end{equation}
where $W\in \mathbb{R}^{(d_{rcnn}+d_{loc})\times d_H}$ and $b\in \mathbb{R}^{d_H}$ are the trainable linear combination weights and bias. The concatenated embeddings $a$ are then passed into a Transformer encoder $T_{image}$, and the final image feature is the hidden output of it:
\begin{equation}
    H_{image} = T_{image}(a)
\end{equation}
where $H_{image} \in \mathbb{R}^{(n+k)\times d_H}$ is the final contextualized representation for one image.

\subsubsection{Scoring Function}
Given the visual and query representations, the matching score can now be computed between a query and an image. Different from other dual-encoder based interaction model, we adopt the fine-grained interaction model proposed by~\citet{zhao2020sparta} to compute the relevance score by:
\begin{align}
    y_i &= \text{max}_{j \in [1, n+k]} (\hat{w}_i^T H_j)  \label{eq:term_match}\\ 
    \phi(y_i) &= \text{ReLU}(y_i + b) \label{eq:sparse}\\
    f(q, v) &= \sum_{i=1}^{m}\log(\phi(y_i)+1)  \label{eq:final}&
\end{align}
where Eq.\ref{eq:term_match} captures the fragment-level interaction between every image region and every query word token; Eq.\ref{eq:sparse} produces sparse embedding outputs via a combination of ReLU and trainable bias, and Eq.\ref{eq:final} sums up the score and prevents an overly large score using $\log$ operation.

\subsection{Retriever training}
Following the training method presented in~\citet{zhao2020sparta}, we use cross entropy loss to train VisualSparta. Concretely, we maximize the objective in Eq.~\ref{eq:loss}, which tries to decide between the ground truth image $v^+$ and irrelevant/random images $V^-$ for each text query $q$. The parameters to learn include both the query encoder $E_{tok}$ and the image transformer encoder $T_{image}$. Parameters are optimized using Adam~\cite{kingma2014adam}.
\begin{equation}
    J = f(q, v^+) - \log\sum_{k \in V^-}e^{f(q, k))}
\label{eq:loss}
\end{equation}
In order to achieve efficient training, we use other image samples from the same batch as negative examples for each training data, an effective technique that is widely used in response selection~\cite{zhang2018personalizing,henderson2019convert}. Preliminary experiments found that as long as the batch size is large enough (we choose to use batch size of 160), this simple approach performs equally well compared to other more sophisticated methods, for example, sample similar images that have nearby labels.

\subsection{Efficient Indexing and Inference}
\label{method:eff-index}
VisualSparta model structure is suitable for real-time inference. As discussed in section~\ref{sec:query-rep}, since query embeddings are non-contextualized, we are able to compute the relationship between each query term $w_i$ and every image $v$ offline. 

Concretely, during offline indexing, for each image $v$, we first compute fragment-level interaction between its regions and every query term in the vocabulary, same as in Eq.~\ref{eq:term_match}. Then, we cache the computed ranking score:
\begin{equation}
    \text{CACHE}(w, v) = \text{Eq.~\ref{eq:sparse}} \label{eq:index}
\end{equation}

During test time, given a query $q=[w_1, ..., w_{m}]$, the ranking score between $q$ and an image $v$ is:
\begin{align}
    &f(q, v) = \sum_{i=1}^{m}\log(\text{CACHE}(w_i, v)+1) \label{eq:infernece}&
\end{align}

As shown in Eq.~\ref{eq:infernece}, the final ranking score during inference time is an O(1) look-up operation followed by summation. Also, the query-time computation can be fit into an Inverted Index architecture~\cite{manning2008introduction}, which enables us to use VisualSparta index with off-the-shelf search engines, for example, Elasticsearch~\cite{gheorghe2015elasticsearch}. 





%% file: experiments.tex



\begin{table*}[!ht]
\centering
\scalebox{0.88}{
\begin{tabular}{l|llllllllll}\hline
 \multicolumn{2}{c}{}& \multicolumn{3}{c}{MSCOCO-1K}&  \multicolumn{3}{c}{MSCOCO-5K}& \multicolumn{3}{c}{Flickr 30K} \\ 

 \multicolumn{2}{c}{} & R@1 & R@5 & R@10 & R@1 & R@5 & R@10 & R@1 & R@5 & R@10 \\ \hline 
 
 \multirow{2}{1.5cm}{Query-dependent} & Unicoder-VL~\cite{li2020unicoder} & 69.7 & 93.5 & 97.2 & 46.7 & 76.0 & 85.3 & 71.5 & 90.9 & 94.9 \\ 
& Oscar ~\cite{li2020oscar} & 75.7 & 95.2 & 98.3 & 54.0 & 80.8 & 88.5 & - & - & - \\\hline

\multirow{8}{1.5cm}{Query-agnostic} & SM-LSTM~\cite{huang2017instance} & 40.7 & 75.8 & 87.4 & - & - & - & 30.2 & 60.4 & 72.3\\
& DAN~\cite{nam2017dual} & - & - & - & - & - & - & 39.4 & 69.2 & 79.1\\
& VSE++~\cite{faghri2017vse++} & 52.0 & - & 92.0 & 30.3 & - & 72.4 & 39.6 & - & 79.5\\
& CAMP~\cite{wang2019camp} & 58.5 & 87.9 & 95.0 & 39.0 & 68.9 & 80.2 & 51.5 & 77.1 & 85.3\\
& SCAN~\cite{lee2018stacked} & 58.8 & 88.4 & 94.8 & 38.6 & 69.3 & 80.4 & 48.6 & 77.7 & 85.2\\
& PFAN~\cite{wang2019position} & 61.6 & 89.6 & 95.2 & - & - & - & 50.4 & 78.7 & 86.1 \\ 
& CVSE~\cite{wang2020consensus} & 59.9 & 89.4 & 95.2 & 35.3 & 66.4 & 78.4 & 52.9 & 80.4 & 87.8 \\

& \textbf{VisualSparta (ours)} & \textbf{68.7} & \textbf{91.2} & \textbf{96.2} & \textbf{45.1} & \textbf{73.0} & \textbf{82.5} &  \textbf{57.1} & \textbf{82.6} & \textbf{88.2} \\ \hline

\end{tabular}}
\caption{Detailed comparisons of text-to-image retrieval results in MSCOCO (1K/5K) and Flickr30K datasets}
\label{tbl:visualsparta-score}

\end{table*}

\subsection{Datasets}
In this paper, we use MSCOCO~\cite{lin2014microsoft}\footnote{\url{https://cocodataset.org}} and Flickr30K~\cite{plummer2015flickr30k}\footnote{\url{http://bryanplummer.com/Flickr30kEntities}} datasets for the training and evaluation of text-to-image retrieval tasks. MSCOCO is a large-scale multi-task dataset including object detection, semantic segmentation, and image captioning data. In this experiment, we follow the previous work and use the image captioning data split for text-to-image model training and evaluation. Following the experimental settings from~\citet{karpathy2015deep}, we split the data into 113,287 images for training, 5,000 images for validation, and 5,000 images for testing. Each image is paired with 5 different captions. The performance of 1,000 (1K) and 5,000 (5K) test splits are reported and compared with previous results.

Flickr30K~\cite{plummer2015flickr30k} is another publicly available image captioning dataset, which contains 31,783 images in total. Following the split from ~\citet{karpathy2015deep}, 29,783 images are used for training, and 1,000 images are used for validation. Scores are reported based on results from 1,000 test images. 

For speed experiments, in addition to MSCOCO 1K and 5K splits, we create 113K split and 1M split, two new data splits to test the performance in the large-scale retrieval setting. Since these splits are only used for speed experiments, we directly reuse the training data from the existing dataset without the concern of data leaking between training and testing phases. Specifically, the 113K split refers to the MSCOCO training set, which contains 113,287 images, $\sim$23 times larger than the MSCOCO 5K test set. The 1M split consists of one million images randomly sampled from the MSCOCO training set. Speed experiments are done on these four splits to give comprehensive comparisons under different sizes of image index.

\subsection{Evaluation Metrics}
\label{sec:visualsparta-metrics}
Following previous works, we use recall rate as our accuracy evaluation metrics. In both MSCOCO and Flikr30K datasets, we report Recall@\textit{t}, \textit{t}=[1, 5, 10] and compare with previous works. 

For speed performance evaluation, we choose query per second and latency(ms) as the evaluation metric to test how each model performs in terms of speed under different sizes of image index.

\subsection{Implementation Details}
All experiments are done using the PyTorch library. During training, one NVIDIA Titan X GPU is used. During speed performance evaluation, one NVIDIA Titan X GPU is used for models that need GPU acceleration. One 10-core Intel 9820X CPU is used for models that needs CPU acceleration. For the image encoder, we initialize the model weights from Oscar-base model~\cite{li2020oscar} with 12 layers, 768 hidden dimensions, and 110M parameters. For the query embedding, we initialize it from the Oscar-base token embedding. The Adam optimizer~\cite{kingma2014adam} is used with the learning rate set to 5e-5. The number of training epochs is set to 20. The input sequence length is set to 120, with 70 for labels with attributes features and 50 for deep visual features. We search on batch sizes (96, 128, 160) with Recall@1 validation accuracy, and set the batch size to 160.

\subsection{Experimental Results}
We compare both recall and speed performance with the current state-of-the-art retrieval model in text-to-image search. Query-dependent model refers to models in which image information cannot be encoded offline, because each image encoding is dependent on the query information. These models usually achieve promising performance in recall but suffer from prohibitively slow inference speed. Query-agnostic model refers to models in which image information can be encoded offline and is independent of query information. In section~\ref{sec:recall-perf} and ~\ref{sec:speed-perf}, we evaluate accuracy and speed performance respectively for both lines of methods.


\subsubsection{Recall Performance}
\label{sec:recall-perf}

As shown in Table~\ref{tbl:visualsparta-score}, the results reveal that our model is competitive compared with previous methods. Among query-agnostic methods, our model is significantly superior to the state-of-the-art results in all evaluation metrics over both MSCOCO and Flickr30K datasets and outperforms previous methods by a large margin. Specifically, in MSCOCO 1K test set, our model outperforms the previously best query-agnostic method~\cite{wang2019position} by 7.1\%, 1.6\%, 1.0\% for Recall@1, 5, 10 respectively. In Flickr30K dataset, VisualSparta also shows strong improvement compared with the previous best method: in Recall@1,5,10, our model gets 4.2\%, 2.2\%, 0.4\% improvement respectively.

We also observe that VisualSparta reduces the gap by a large margin between query-agnostic and query-dependent methods. In MSCOCO-1K split, the performance of VisualSparta is only 1.0\%, 2.3\%, 1.0\% lower than Unicoder-VL method~\cite{li2020unicoder} for Recall@1,5,10 respectively. Compared to Oscar~\cite{li2020oscar}, the current state-of-the-art query-dependent model, our model is 7\% lower than the Oscar model in MSCOCO-1K Recall@1. This shows that there is still room for improvement in terms of accuracy for query-agnostic model.



\subsubsection{Speed Performance}
\label{sec:speed-perf}
\begin{table}[ht]
\centering
\scalebox{0.9}{
\begin{tabular}{P{0.16\textwidth}p{0.06\textwidth}p{0.06\textwidth}p{0.06\textwidth}p{0.06\textwidth}}
 

& \multicolumn{2}{c}{GPU}& \multicolumn{2}{c}{CPU} \\ 
\cmidrule(lr){2-3}\cmidrule(lr){4-5}

Index Size vs. Query/s & Oscar & CVSE & CVSE & Visual Sparta \\ \hline

1K        & 0.4    & 195.1 & 177.4& \textbf{451.4}        \\
5K        & 0.06   & 191.0 & 162.0& \textbf{390.5}        \\
113K      & 0.003  & 101.2 & 5.4   & \textbf{275.5}        \\
1M        & 0.0003 & 21.7 & 0.3   & \textbf{117.3}        \\ \hline
\end{tabular}}
\caption{\textbf{Model Speed vs. Index Size}: VisualSparta experiments are done under setting top-\textit{n} term scores to 1000. Detailed settings are reported in section \ref{sec:speed-perf}.}
\label{tbl:visualsparta-speed}
\end{table}
To show the efficiency of VisualSparta model in both small-scale and large-scale settings, we create 113K dataset and 1M dataset in addition to the original 1K and 5K test split, as discussed in section~\ref{sec:visualsparta-metrics}. Speed experiments are done using these four splits as testbeds.

\begin{table*}[ht!]
\centering
\small
\begin{tabular}{ccc@{\hskip 0.6cm}ccc@{\hskip 0.6cm}ccc}\hline
 & & & \multicolumn{3}{c}{MSCOCO-1k}&  \multicolumn{3}{c}{MSCOCO-5k} \\ \hline 

 \textit{n} & Inf. time (ms)$\downarrow$ & query/s$\uparrow$ & R@1$\uparrow$ & R@5$\uparrow$ & R@10$\uparrow$ & R@1$\uparrow$ & R@5$\uparrow$ & R@10$\uparrow$ \\ \hline  

50 & 1.9 & 537.0 & 54.6	& 82.8	& 90.7	& 33.0	& 60.0	& 71.1\\
100	& 1.9 & 514.7	& 60.1	& 86.2	& 92.8	& 37.1	& 64.6	& 75.3\\
500	& 2.1 & 477.7 & 65.5	& 90.3	& 95.1	& 42.5	& 70.6	& 80.4\\
1000 & 2.4 & 414.5 & 67.5	& 90.9	& 95.8	& 43.7	& 71.7	& 81.5\\
2000 & 3.9 & 256.3 & 68.5	& 91.1	& 96.0	& 44.4	& 72.5	& 82.1\\
all & 6.9 & 144.1 & 68.7	& 91.2	& 96.2	& 45.1	& 73.0	& 82.5\\ \hline

\end{tabular}
\caption{Effect of top-\textit{n} term scores in terms of speed and accuracy tested in MSCOCO dataset; $\uparrow$ means higher the better, and $\downarrow$ means lower the better.}
\label{tbl:visualsparta-tscore}
\end{table*}

To make a fair comparison, we benchmark each method with its preferred hardware and software for speed acceleration. Specifically, For CVSE model~\cite{wang2020consensus}, both CPU and GPU inference time are recorded. For CPU setting, the Maximum Inner Product Search (MIPS) is performed using their original code based on Numpy~\cite{harris2020array}. For GPU setting, we adopt the model and use FAISS~\cite{johnson2019billion}, an optimized MIPS library, to test the speed performance. For Oscar model~\cite{li2020oscar}, since the query-dependent method cannot be formulated as a MIPS problem, we run the original model using GPU acceleration and record the speed. For VisualSparta, we use the top-1000 term scores settings for the experiment. Since VisualSparta can be fit into an inverted-index architecture, GPU acceleration is not required. For all experiments, we use 5000 queries from MSCOCO-1K split as query input to test the speed performance.


As we can see from Table~\ref{tbl:visualsparta-speed}, in all four data splits (1K, 5K, 113K, 1M), VisualSparta significantly outperforms both the best query-agnostic model (CVSE~\cite{wang2020consensus}) and the best query-dependent model (Oscar~\cite{li2020oscar}). Under CPU comparison, the speed of VisualSparta is 2.5, 2.4, 51, and 391 times faster than that of the CVSE model in 1K, 5K, 113K, and 1M splits respectively.

This speed advantage also holds even if previous models are accelerated with GPU acceleration. To apply the latest MIPS progress to the comparison, we adopt the CVSE model to use FAISS~\cite{johnson2019billion} for better speed acceleration. Results in the table reveal that the speed of VisualSparta can also beat that of CVSE by 2.5X in the 1K setting, and this speed advantage increases to 5.4X when the index size increases to 1M. 

Our model holds an absolute advantage when comparing speed to query-dependent models such as Oscar~\cite{li2020oscar}. Since the image encoding is dependent on the query information, no offline indexing can be done for the query-dependent model. As shown in Table~\ref{tbl:visualsparta-speed}, even with GPU acceleration, Oscar model is prohibitively slow: In the 1K setting, Oscar is $\sim$1128 times slower than VisualSparta. The number increases to 391,000 when index size increases to 1M.




%% file: analysis.tex

\subsection{Speed-Accuracy Flexibility}
\label{sec:speed-accu-flex}

As described in section~\ref{method:eff-index}, each image can be well represented by a list of weighted tokens independently. This feature makes VisualSparta flexible during indexing time: users can choose to index using top-\textit{n} term scores based on their memory constraint or speed requirement. 

Table~\ref{tbl:visualsparta-tscore} compares recall and speed in both MSCOCO 1K and 5K split under different choices of \textit{n}. From the comparison between using all term scores and using top-2000 term scores, we found that VisualSparta can get $\sim$1.8X speedup with almost no performance drop. if higher speed is needed, \textit{n} can always be set to a lower number with a sacrifice of accuracy, as shown in Table~\ref{tbl:visualsparta-tscore}.

Figure~\ref{fig:visualsparta-r1} visualizes the trade-off between model accuracy and inference speed. The x-axis represents the average inference time of a single query in millisecond, and the y-axis denotes the Recall@1 on MSCOCO 1K test set. For VisualSparta, each dot represents the model performance under certain top-\textit{n} term score settings. For other methods, each dot represents their speed and accuracy performance. The curve reveals that with larger n, the recall becomes higher and the speed gets slower. From the comparison between VisualSparta and other methods, we observe that by setting top-\textit{n} term scores to 500, VisualSparta can already beat the accuracy performance of both PFAN~\cite{wang2019position} and CVSE~\cite{wang2020consensus} with $\sim$2.8X speedup.

\subsection{Ablation Study on Image Encoder}

\begin{table*}[!ht]
\centering
\small
\scalebox{1.0}{
\begin{tabular}{llllllll}\hline
& & \multicolumn{3}{c}{MSCOCO-1k}&  \multicolumn{3}{c}{MSCOCO-5k}\\ 

\# & & R@1 & R@5 & R@10 & R@1 & R@5 & R@10 \\ \hline

1&VisualSparta & \textbf{68.7} & 91.2 & 96.2 & \textbf{45.1} & \textbf{73.0} & \textbf{82.5} \\ \hline

2&\hspace{0.2cm}$-$ attributes features & 68.2(-0.5) & \textbf{91.8}(+0.6) & \textbf{96.3}(+0.1) & 44.4(-0.7) & 72.8(-0.2) & 82.4(-0.1) \\

3&\hspace{0.2cm}$-$ labels w. attributes features & 66.7(-2.0) & 91.2(+0.0) & 95.9(-0.3) & 43.4(-1.7) & 71.6(-1.4) & 81.6(-0.9) \\

4&\hspace{0.2cm}$-$ visual features & 49.1(-19.6) & 80.3(-10.9) & 89.4(-6.8) & 26.5(-18.6) & 54.1(-18.9) & 66.8(-15.7) \\\hline


\end{tabular}}
\caption{Ablation study with using different features in the image answer encoding}
\label{tbl:visualsparta-ablation}
\end{table*}

\begin{figure*}[!ht]
\centering
\includegraphics[width=14.8cm]{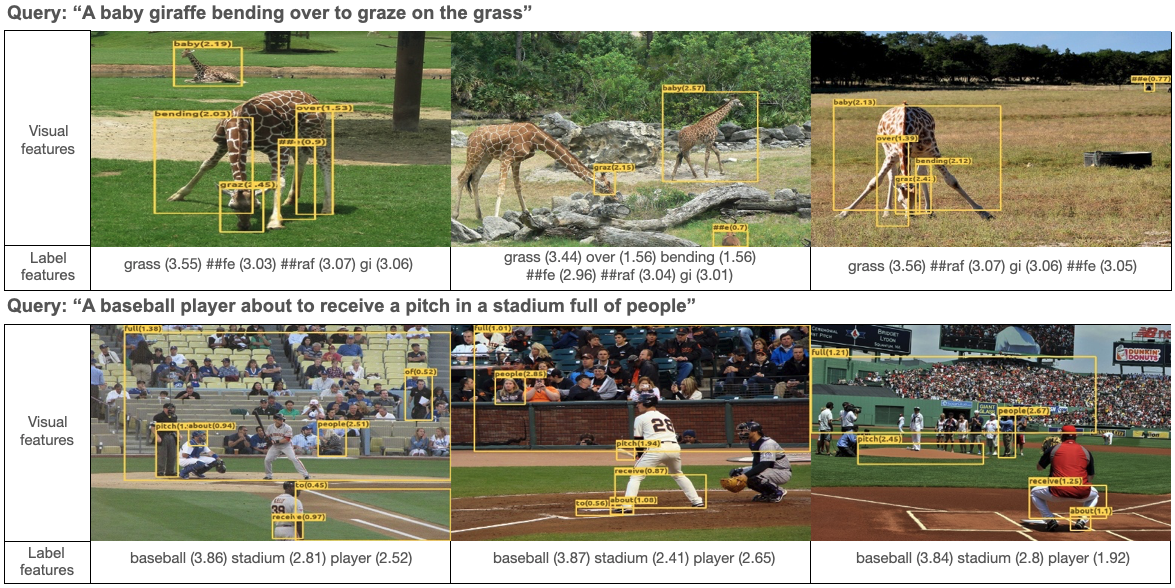}
\caption{Example retrieved images with features attended given query terms; term scores are in parentheses.} 
\label{fig:visualsparta-example}
\end{figure*}

As shown in Figure~\ref{fig:visualsparta-model}, the image encoder takes a concatenation of object label features with attributes and deep visual features as input. In this section, we do an ablation study and analyze the contributions of each part of the image features to the final score. 

In Table \ref{tbl:visualsparta-ablation}, different components are removed from the image encoder for performance comparison. From the table, we observe that removing either attributes features (row 1) or label features with attributes (row 2) only hurts the performance by a small margin. However, when dropping visual features and only using label with attributes features for image representation (row 3), it appears that the model performance drops by a large margin, where the Recall@1 score drops from 68.7\% to 49.1\%($-$19.6\%).

From this ablation study, we can conclude that deep visual features make the most contribution to the VisualSparta model structure, which shows that deep visual features are significantly more expressive compared to textual features, i.e., label with attributes features. More importantly, it shows that VisualSparta is capable of learning cross-modal knowledge, and the biggest gain indeed comes from learning to match query term embeddings with deep visual representations.

\subsection{Cross-domain Generalization}

\begin{table}[!ht]
\centering
\scalebox{0.85}{
\begin{tabular}{p{0.3\textwidth}|p{0.05\textwidth}p{0.05\textwidth}p{0.05\textwidth}} \hline
Models & R@1  & R@5  & R@10 \\ \hline

VSE++\cite{faghri2017vse++} & 28.4 & 55.4 & 66.6 \\
LVSE\cite{engilberge2018finding} & 34.9   & 62.4 & 73.5 \\
SCAN\cite{lee2018stacked} & 38.4  & 65.0   & 74.4 \\
CVSE\cite{wang2020consensus} & 38.9 & 67.3   & 76.1 \\ \hline
\textbf{VisualSparta (ours)} & \textbf{45.4} & \textbf{71.0} & \textbf{79.2} \\ \hline

\end{tabular}}

\caption{Cross-dataset performance; models are trained on MSCOCO dataset and tested on Flickr30K dataset. }
\label{tbl:visualsparta-crossdomain}
\end{table}
Table \ref{tbl:visualsparta-crossdomain} shows the cross-domain performance for different models. All models are trained on MSCOCO and tested on Flickr30K. We can see from the table that VisualSparta consistently outperforms other models in this setting. This indicates that the performance of VisualSparta is consistent across different data distributions, and the performance gain compared to other models is also consistent when testing in this cross-dataset settings.  


\subsection{Qualitative Examples}
We query VisualSparta on the MSOCO 113K split and check the results.  As shown in Figure~\ref{fig:visualsparta-example}, visual and label features together represent the max attended features for given query tokens. Interestingly, we observe that VisualSparta model is capable of grounding adjectives and verbs to the relevant image regions. For example, ``graz" grounds to the head of giraffe in the first example. This further confirms the hypothesis that weighted bag-of-words is a valid and rich representation for images.


%% file: conclusion.tex
In conclusion, this paper presents VisualSparta, an accurate and efficient text-to-image retrieval model that shows the state-of-the-art scalable performance in both MSCOCO and Flickr30K. 
Its main novelty lies in the combination of powerful pre-trained image encoder with fragment-level scoring. Detailed analysis also demonstrates that our approach has substantial scalability advantages compared to previous best methods when indexing large image datasets for real-time searching, making it suitable for real-world deployment.